\documentclass[11pt]{article}
\usepackage[margin=1in]{geometry}
\usepackage{amsmath,amssymb,amsthm,mathtools,booktabs}
\usepackage[numbers,sort&compress]{natbib}
\usepackage{hyperref}
\newtheorem{definition}{Definition}[section]
\newtheorem{assumption}[definition]{Assumption}

\newtheorem{proposition}[definition]{Proposition}
\newtheorem{theorem}[definition]{Theorem}
\newtheorem{corollary}[definition]{Corollary}
\newcommand{\R}{\mathbb{R}}
\newcommand{\F}{\mathbb{F}}
\DeclareMathOperator{\Dgm}{Dgm}
\newcommand{\FinRel}{\mathbf{FinRel}}
\newcommand{\StrState}{\mathsf{StrState}}
\newcommand{\StrTrack}{\mathsf{StrTrack}}
\newcommand{\Dstr}{\Delta_{\mathrm{str}}}
\newcommand{\id}{\mathrm{id}}
\newcommand{\dom}{\operatorname{dom}}
\newcommand{\im}{\operatorname{im}}
\title{Structural Imagination Spaces: A Formal Theory of Layered Internal Structure}
\author{Chulsu Ha\\Independent Researcher\\\texttt{charles.andrew.ha@gmail.com}}
\date{20 August 2026}
\begin{document}
\maketitle
\begin{abstract}
The Theory of Structural Imagination (TSI) proposes that an internal
representation for reasoning and action should be specified by the structures it
preserves and the transformations it supports, rather than by its storage format
alone. This paper gives a consolidated formal core. A state is a typed
relational functor on a finite presented schema, a simplicial complex and metric
on a common tagged carrier, a monotone filtration, a full-support mass, and
action-conditioned tracked transitions. We define structural isomorphism and
five-layer survivor preservation, including the cross-layer compatibility axiom,
under which we prove a quantitative capacity bound on the relational defect. We
also prove equivalence and composition properties, a six-component zero criterion
including turnover, a hard-label metric--measure triangle inequality, and a
quotient-level rollout bound under explicit equivariance and Lipschitz assumptions. These are
conditional mathematical results about the specified representation class; they
do not establish that human cognition or an unconstrained neural network
implements it.
\end{abstract}

\section{Introduction}

Representation learning routinely stores internal states as vectors or tensors.
That storage choice does not specify which entities persist, which relations
compose, which metric quantities matter, how ordered and mass observables behave,
or how an action changes the state. TSI begins with a narrower question: what
mathematical data must be exposed if an internal state is intended to support
structural simulation, reasoning, and action?

The formal object developed here is a finite presented schema together with a
common tagged carrier, typed relations, a simplicial complex, a metric, a
filtration, a full-support mass, and action-conditioned tracked transitions.
These layers are allowed to interact, but they are not conflated. A vector may
encode the data; its coordinates are not the complete structural specification.

The paper makes one integrated contribution through five components:
(1) a state contract with a declared cross-layer compatibility axiom;
(2) an exact structural isomorphism and its equivalence proof;
(3) survivor-level preservation for five observables and its composition proof;
(4) a six-defect zero criterion separating local preservation from turnover; and
(5) categorical and quotient-level extensions, including metric--measure
comparison and rollout stability.

The finite-relation category and presented-category descent are imported from
the calculus of relations and categorical data formalisms
\cite{freydscedrov1990categories,spivak2012functorial,
spivakwisnesky2015relational}. Filtered persistence stability is imported from
persistent homology \cite{cohensteiner2007stability,chazal2009proximity};
metric--measure couplings and gluing are imported from Gromov--Wasserstein and
optimal-transport theory \cite{memoli2011gromov,peyre2019computational}; and the
rollout recurrence uses the standard Lipschitz error-propagation pattern
\cite{asadi2018lipschitz}. The TSI-specific contribution is their integration
under a common tagged state contract, the compatibility invariant, the
label-preserving partial transition category, and the explicit separation of
survivor preservation from temporal identity. The paper does not claim that
these layers are necessary or sufficient for intelligence, that they describe
human mathematical imagery, or that a neural model will discover them without
inductive constraints.

\section{Related Work and Positioning}

Structured learning systems commonly impose object- and relation-centered
inductive biases through interaction networks, graph networks, or structured
world models \cite{battaglia2016interaction,battaglia2018relational,
kipf2020contrastive}. These works specify learnable architectures and training
objectives. TSI instead specifies a mathematical state contract, its
equivalences, and conditional preservation properties. It neither supplies a
neural discovery algorithm nor implies that the contract will emerge from an
unconstrained learner.

Categorical database models treat a schema as a small category and an instance
as a set-valued functor \cite{maclane1998categories,spivak2012functorial};
finitely presented schemas
can be given by directed multigraphs and path equations
\cite{spivakwisnesky2015relational}. TSI imports this presentation discipline
but uses finite binary relations as morphisms, following the standard calculus
of relations \cite{freydscedrov1990categories}. Its additional simplicial,
metric, filtration, mass, and tracked-transition layers are not part of those
database formalisms.

The topological extension uses persistence diagrams and their stability theory
\cite{edelsbrunner2002topological,zomorodian2005computing,
cohensteiner2007stability,chazal2009proximity,bauerlesnick2015induced,
crawleyboevey2015decomposition}. The metric--measure extension is
a finite, hard-label-constrained variant of a Gromov--Wasserstein comparison
\cite{memoli2011gromov}; it is distinct from formulations that jointly compare
features and structure \cite{vayer2020fused}. Standard coupling and gluing
arguments are drawn from optimal transport
\cite{villani2009optimal,peyre2019computational}.

Bisimulation metrics for Markov decision processes build behavioral distances
from rewards and transition kernels \cite{ferns2004metrics}. Multi-step model
error has also been bounded under Lipschitz assumptions in Wasserstein geometry
\cite{asadi2018lipschitz}, while model-based policy optimization makes the
practical case for restricting model-generated rollout horizons
\cite{janner2019trust}. TSI does not identify its structural discrepancy with
a bisimulation fixed point or a reward-sufficient abstraction. Its rollout
result is the deterministic error recurrence that follows once a quotient-level
extended pseudometric, equivariant updates, one-step errors, and Lipschitz
constants have separately been supplied.

\section{Formal Framework}

\subsection{Typed relational carriers}

We use the category $\FinRel$ of finite sets and binary relations. Its objects are
finite sets, and a morphism $R:X\nrightarrow Y$ is a subset of $X\times Y$.
For $R:X\nrightarrow Y$ and $S:Y\nrightarrow Z$,
\[
S\circ R=\{(x,z):\exists y\in Y,
(x,y)\in R,\ (y,z)\in S\}.
\]
The identity on $X$ is $\Delta_X=\{(x,x):x\in X\}$.

Fix a finite quiver $Q=(Q_0,Q_1,s,t)$ and a finite set $\mathcal E$ of
pairs of parallel paths in its free path category. The category
$\operatorname{Path}(Q)$ has the vertices of $Q$ as objects, finite composable
paths as morphisms, length-zero paths as identities, and concatenation as
composition. The schema is the presented
category
\[
\mathcal S=\operatorname{Path}(Q)/\!\equiv_{\mathcal E},
\]
where $\equiv_{\mathcal E}$ is the smallest path congruence containing
$\mathcal E$. The arrows of $Q$ are the declared generators, and every schema
morphism has a path representative subject to the declared equations. A typed
relational carrier is a functor $F:\mathcal S\to\FinRel$ such that every
$F(A)$ is nonempty and finite. A generator $e\in Q_1$ is interpreted as the
relation $F([e])$. Functoriality requires
$F(\id_A)=\Delta_{F(A)}$ and $F(g\circ f)=F(g)\circ F(f)$.
A label family is a collection of maps
$\lambda_A:F(A)\to L_A$ into fixed typewise label sets.

\begin{definition}[Integrated structural imagination state]
An integrated structural imagination state is a tuple
\[
\Sigma=(F,\lambda,K,d,f,\mu),
\]
where $F$ is a typed relational carrier, $\lambda$ is a label family,
$K$ is a finite abstract simplicial complex whose vertex set is the tagged union
$|F|=\coprod_A F(A)$, and $d$ is a metric on $|F|$. The function
$f:K\to\mathbb R$ is a monotone filtration and
$\mu:|F|\to(0,1]$ is a full-support probability mass function. The state also
obeys the compatibility condition
\[
(x,y)\in F([e])\Longrightarrow\{x,y\}\in K\qquad(e\in Q_1).
\]
The class of such states is denoted by $\StrState(Q,\mathcal E)$.
\end{definition}

The common tagged carrier is a scope decision. It permits relations, simplices,
metric distances, filtration values, and mass to refer to the same typed
entities. The compatibility condition is the minimal cross-layer constraint
used in this paper: a declared generator relation must be visible as a
one-simplex, but no converse or metric compatibility is assumed.

\begin{definition}[Integrated structural isomorphism]
Let $\Sigma=(F,\lambda,K,d,f,\mu)$ and
$\Sigma'=(G,\nu,L,\rho,g,\eta)$. An integrated structural isomorphism
$\varphi:\Sigma\to\Sigma'$ is a family of bijections
$\varphi_A:F(A)\to G(A)$ such that:
\begin{enumerate}
\item $\nu_A(\varphi_Ax)=\lambda_A(x)$ for every $A,x$;
\item the tagged union $\bar\varphi:|F|\to|G|$ satisfies
$\bar\varphi[K]=L$;
\item $\rho(\bar\varphi x,\bar\varphi y)=d(x,y)$ for all $x,y$;
\item $g(\bar\varphi\sigma)=f(\sigma)$ for every $\sigma\in K$;
\item $\eta(\bar\varphi x)=\mu(x)$ for every $x\in|F|$;
\item for every generator $e\in Q_1$,
\[
((\varphi_{s(e)}\times\varphi_{t(e)})[F([e])])=G([e]).
\]
\end{enumerate}
We write $\Sigma\cong_{\mathrm{str}}\Sigma'$ when such a family exists.
\end{definition}

\begin{definition}[Tracked transition]
A tracked transition $p:\Sigma\rightharpoonup\Sigma'$ is a family
$p=(p_A)_A$ in which each $p_A$ is an injective map from a subset of $F(A)$
into $G(A)$ and
\[
\nu_A(p_Ax)=\lambda_A(x)\qquad(x\in\dom p_A).
\]
Its domain and image tagged unions are written $D_p$ and $I_p$. The identity is
the total identity family. If
$p:\Sigma_0\rightharpoonup\Sigma_1$ and
$q:\Sigma_1\rightharpoonup\Sigma_2$, their composite is the family
$(q_A\circ p_A)|_{p_A^{-1}(\dom q_A)}$.
\end{definition}

\begin{definition}[Tagged extension of a partial transition]
For a tracked transition $p=(p_A)_A$, its tagged extension $\bar p$ is the
map on $D_p=\coprod_A\dom(p_A)$ defined by $\bar p(x)=p_A(x)$ when
$x\in F(A)$. Its image is $I_p=\coprod_A\im(p_A)$.
\end{definition}

\begin{proposition}[The category of tracked transitions]
Integrated states and tracked transitions form a category
$\StrTrack(Q,\mathcal E)$.
\end{proposition}

\begin{proof}
The identity family is injective, label-preserving, and has the full carrier as
its domain. A composite is injective on the restricted domain. If $x$ is in
its domain, label preservation follows successively from $p$ and $q$. The
domain of a triple composite is the set of elements on which all three partial
maps are defined, and both parenthesizations have the same value there. Hence
composition is associative. Restricting either composite by an identity leaves
the same partial family, proving the unit laws.
\end{proof}

\begin{definition}[Action-conditioned transition]
Let $\mathcal A$ be an action set. An action-conditioned transition is a pair
$(a,p)$ with $a\in\mathcal A$ and $p$ a tracked transition. The action label is
an index on the transition, not additional data in the morphism of
$\StrTrack(Q,\mathcal E)$; consequently composition is defined on the tracked
maps and does not concatenate incompatible single labels.
\end{definition}

\section{Running Example: A Four-Surface Star State}

We give one finite state that satisfies all state axioms and can be reused across
the extensions. Let the presented schema have two objects $B$ (block) and
$S$ (surface), one generator $e:B\to S$, and no nontrivial path equations.
Set
\[
F(B)=\{b\},\qquad F(S)=\{s_1,s_2,s_3,s_4\},
\]
and let $\lambda_B(b)=\mathsf{block}$ and $\lambda_S(s_i)=\mathsf{surface}$ for all $i$. Let $F([e])=\{(b,s_i):1\leq i\leq4\}$. Let $K$ be the star complex whose
vertices are these five entities and whose nontrivial simplices are the four
edges $\{b,s_i\}$. The compatibility condition holds by construction.

Define
\[
d(x,x)=0,\qquad d(b,s_i)=1,\qquad d(s_i,s_j)=2\quad(i\ne j),
\]
with symmetry understood. This is a metric: the only nontrivial triangle
inequality is $2\leq1+1$. Give every vertex filtration value $0$, every edge
filtration value $1$, and every entity mass $1/5$. Call the resulting state
$\Sigma_\star$. The filtration is monotone and the mass has full support.

Let $\Sigma'_\star$ be an isomorphic copy with primed entity names. The
bijection $b\mapsto b'$ and $s_i\mapsto s_i'$ preserves labels, the star
complex, distances, filtration values, masses, and the generator relation.
Consequently,
\[
d_B(\Dgm_k(f),\Dgm_k(g))=0\quad(k\leq r),\qquad
\Delta_{\mathrm{mm},p}((|F|,d,\mu,\lambda),(|G|,\rho,\eta,\nu))=0
\quad(1\leq p<\infty).
\]
The degree-zero persistence diagram has one point $(0,\infty)$ and four points $(0,1)$, one for each component born before the four edges appear; there is no
positive-degree class for this star filtration.

The same example also exposes the distinction between local preservation and
temporal identity. Let $p$ map $b,s_1,s_2$ to $b',s_1',s_2'$ and leave
$s_3,s_4$ unmatched. The survivor subcomplexes and restricted generator
relations agree, all surviving distances, filtration values, and masses agree,
so
\[
m_K(p)=m_d(p)=m_R(p)=m_f(p)=m_\mu(p)=0.
\]
However,
\[
b(p)=(4-2)+(4-2)=4,
\]
where the first term is source turnover in type $S$ and the second is target
turnover in type $S$. Thus zero layer defects do not certify full temporal
identity. If instead all four surfaces and the block are matched by the primed
copy, then all six quantities, including $b(p)$, vanish and the transition is
a full structural isomorphism.

This example is deliberately finite and transparent. It does not claim that
the star complex is a model of cognition; it verifies that the formal state
class is nonempty and that the defect quantities distinguish survivor-level
preservation from carrier turnover.
\section{Layer Theorems}

\begin{proposition}[Structural equivalence]
The relation $\cong_{\mathrm{str}}$ on $\StrState(Q,\mathcal E)$ is reflexive,
symmetric, and transitive.
\end{proposition}

\begin{proof}
The identity family preserves labels, simplices, distances, filtration values,
masses, and generator relations, so reflexivity holds. If $\varphi$ is an
isomorphism, each $\varphi_A^{-1}$ is a bijection; applying inverse maps to all
six defining equalities gives the reverse conditions, so symmetry holds. If
$\varphi$ and $\psi$ are composable, their typewise composites preserve labels.
The simplex, metric, filtration, mass, and generator-relation equalities compose
in sequence, giving all six conditions for $\psi\varphi$. Thus transitivity
holds.
\end{proof}

\begin{proposition}[Generator preservation extends to paths]
If an integrated structural isomorphism preserves every generating relation,
then it transports the relation assigned to every morphism of $\mathcal S$.
\end{proposition}

\begin{proof}
For an identity morphism, a bijection transports the diagonal relation to a
diagonal relation. Suppose the claim holds for paths $u:A\to B$ and $v:B\to C$.
Membership in $F(v)\circ F(u)$ is witnessed by an intermediate $y\in F(B)$.
The induction hypotheses transport both relations, and bijectivity of
$\varphi_B$ transports the witness to and from every element of $G(B)$. Hence
the transported composite is $G(v)\circ G(u)$. Induction on path length,
followed by the path equations, proves the claim for every schema morphism.
\end{proof}

For a tracked transition $p$, write $K|_{D_p}$ for the induced subcomplex on
surviving tagged entities and $F([e])|_{D_A,D_B}$ for the restriction of a
generator relation to survivor domains.

\begin{definition}[Survivor-level layer preservation]
A transition $p:\Sigma\rightharpoonup\Sigma'$ preserves the following layers on
survivors when
\begin{align*}
\bar p[K|_{D_p}]&=L|_{I_p} &&\text{(topology)},\\
d'(\bar p x,\bar p y)&=d(x,y) &&\text{for all }x,y\in D_p &&\text{(geometry)},\\
(p_A\times p_B)[F([e])|_{D_A,D_B}]&=G([e])|_{I_A,I_B}
 &&\text{for every generator }e:A\to B &&\text{(relations)},\\
g(\bar p\sigma)&=f(\sigma)
 &&\text{for every surviving simplex }\sigma &&\text{(filtration)},\\
\eta(\bar p x)&=\mu(x)
 &&\text{for every }x\in D_p &&\text{(mass)}.
\end{align*}
\end{definition}

\begin{theorem}[Composition closure]
Let $\Sigma_0\xrightarrow{p}\Sigma_1\xrightarrow{q}\Sigma_2$ be tracked
transitions. If both transitions preserve any chosen subset of topology,
geometry, relations, filtration, and mass on survivors, then $q\circ p$
preserves the same subset.
\end{theorem}

\begin{proof}
For each type $A$, let
$E_A=\{x\in\dom(p_A):p_Ax\in\dom(q_A)\}$,
$J_A=p_A(E_A)$, and $H_A=q_A(J_A)$. These are the domains, intermediate
survivors, and images of the composite. The first topology and filtration
equalities restricted to $E=\coprod_AE_A$ give the structures on $J$, and the
second equalities restricted to $J$ give the structures on $H$; composing gives
the required equalities. For geometry and mass, applying the two pointwise
equalities successively gives
$d_2(\bar q\bar p x,\bar q\bar p y)=d_0(x,y)$ and
$\eta(\bar q\bar p x)=\mu(x)$. For a generator $e:A\to B$, restrict the first
relational equality to $E_A\times E_B$, whose image is $J_A\times J_B$, and
then restrict the second to $J_A\times J_B$. The product maps are bijections on
the corresponding survivor products, so the relational equality composes.
\end{proof}

\begin{corollary}[Trajectory preservation]
Transitions preserving a chosen subset of these five layers form a wide
subcategory of $\StrTrack(Q,\mathcal E)$. Consequently, a finite trajectory whose
edges preserve those layers has a composite transition preserving them.
\end{corollary}

\begin{proof}
The identity transition preserves every layer. The composition theorem gives
closure under composition, and induction on the number of trajectory edges
gives the result.
\end{proof}

\section{Persistent Topology Extension}

This section adds a persistent topological observable to the integrated state.
It is deliberately narrower than a claim that topology identifies structure:
the carrier is a fixed finite simplicial complex, and the filtration is the
only varying input in the stability result below.

\begin{definition}[Filtered finite carrier]
Let $K$ be a finite abstract simplicial complex and let $\F$ be a field. A
filtration function is a map $f:K\to\R$ satisfying $\sigma\subseteq\tau
\Rightarrow f(\sigma)\leq f(\tau)$. Its sublevel complex at $a\in\R$ is
\[
 K_a^f=\{\sigma\in K:f(\sigma)\leq a\}.
\]
The persistent signature in degree $k$ is the persistence diagram
$\Dgm_k(f)$ of the homology persistence module
$H_k(K_a^f;\F)$, and the truncated signature is
$\Pi_r(K,f)=(\Dgm_0(f),\ldots,\Dgm_d(f))$. This construction follows the
standard finite persistence framework
\cite{edelsbrunner2002topological,zomorodian2005computing}.
\end{definition}

The monotonicity condition makes every $K_a^f$ a simplicial subcomplex: if
$\tau$ is retained, every face $\sigma$ is retained as well. Thus inclusion
of sublevel complexes gives the persistence maps used in the definition.

\begin{definition}[Persistent discrepancy]
For two filtration functions $f,g$ on the same finite carrier $K$, define
\[
 \Delta_{\mathrm{pers},r}(f,g)=\max_{0\leq k\leq r}
 d_B\bigl(\Dgm_k(f),\Dgm_k(g)\bigr),
\]
where $d_B$ is the bottleneck distance. The two filtrations are compared only
after fixing the field, the carrier, the degree cutoff, and the units of the
filtration values.
\end{definition}

The following result is the standard stability theorem for persistence
diagrams and persistence modules
\cite{cohensteiner2007stability,chazal2009proximity}.

\begin{theorem}[Filtered stability; imported theorem]
For a fixed finite simplicial complex $K$ and monotone filtration functions
$f,g:K\to\R$,
\[
 d_B\bigl(\Dgm_k(f),\Dgm_k(g)\bigr)\leq \|f-g\|_\infty
 \quad\text{for every }k.
\]
Consequently, $\Delta_{\mathrm{pers},r}(f,g)\leq\|f-g\|_\infty$.
\end{theorem}

\begin{proof}
Put $\delta=\|f-g\|_\infty$. For every simplex $\sigma$, the inequalities
$g(\sigma)\leq f(\sigma)+\delta$ and
$f(\sigma)\leq g(\sigma)+\delta$ hold. Therefore, for every $a$,
\[
 K_a^f\subseteq K_{a+\delta}^g
 \quad\text{and}\quad
 K_a^g\subseteq K_{a+\delta}^f.
\]
The inclusions commute with the inclusions between sublevel complexes as $a$
increases. Applying homology gives a $\delta$-interleaving of the two
persistence modules in degree $k$. Because $K$ is finite, every vector space
in either module is finite-dimensional; in particular, the modules are
q-tame and admit interval decompositions and persistence diagrams
\cite{crawleyboevey2015decomposition}. The algebraic stability theorem
\cite{chazal2009proximity,bauerlesnick2015induced} gives a bottleneck distance no larger than
the interleaving distance. Hence $d_B(\Dgm_k(f),\Dgm_k(g))\leq\delta$.
Taking the maximum over $0\leq k\leq r$ proves the final inequality. The
only non-elementary step is the imported algebraic stability theorem; the
sublevel inclusions and interleaving construction are established here.
\end{proof}

\begin{proposition}[Isomorphism invariance of the persistent signature]
Let $\varphi:K\to L$ be a simplicial isomorphism and let $f:K\to\R$ and
$g:L\to\R$ satisfy $g(\varphi(\sigma))=f(\sigma)$ for every simplex
$\sigma$. Then, for every $k$,
\[
 d_B\bigl(\Dgm_k(f),\Dgm_k(g)\bigr)=0.
\]
\end{proposition}

\begin{proof}
Fix $a$. The value-preservation condition implies
$\varphi(K_a^f)=L_a^g$. Since the restriction
$\varphi_a:K_a^f\to L_a^g$ is a simplicial isomorphism, functoriality of
simplicial homology gives $H_k(K_a^f;\F)\cong H_k(L_a^g;\F)$. For
$a\leq b$, the square formed by these restrictions and the inclusion-induced
homology maps commutes. Thus the persistence modules are isomorphic and their
persistence diagrams are identical, so the bottleneck distance is zero.
\end{proof}

\begin{proposition}[The persistent signature is not complete for a labeled carrier]
Even on one fixed labeled carrier, equality of all persistence diagrams need
not determine the filtration as labeled structural data.
\end{proposition}

\begin{proof}
Let $K$ consist of two isolated vertices $x,y$ with distinct labels. Set
$f(x)=0,f(y)=1$ and $g(x)=1,g(y)=0$. Both degree-zero diagrams contain
$(0,\infty)$ and $(1,\infty)$, and every positive-degree diagram is empty.
Thus $\Pi_r(K,f)=\Pi_r(K,g)$ for every $r\geq0$. However, the only
label-preserving bijection fixes both vertices, and it does not preserve the
filtration values. Hence the persistent signature does not determine the
labeled filtered state.
\end{proof}

The extension certifies invariance and controlled sensitivity under its stated
hypotheses. It does not establish completeness, comparison of unrelated
carriers without a correspondence, or neural discovery of the carrier. Those
questions require separate assumptions or empirical validation.

\begin{proposition}[State-level persistent invariance]
Let $\Sigma$ and $\Sigma'$ be integrated states and let
$\varphi:\Sigma\to\Sigma'$ be a structural isomorphism. Then for every
degree cutoff $r$,
\[
d_B(\Dgm_k(f),\Dgm_k(g))=0\quad\text{for every }k\leq r.
\]
\end{proposition}

\begin{proof}
The filtration-preserving component of $\varphi$ satisfies
$g(\bar\varphi\sigma)=f(\sigma)$ for every simplex. Therefore the
isomorphism-invariance proposition applies to the two filtered carriers.
\end{proof}

\section{Metric--Measure Extension}

The geometric layer can contain uncertain or repeated entities, so a set-level
correspondence alone is not sufficient. We attach probability mass to the
finite metric carrier and compare mass through couplings. The result below is
an exactness theorem, not a finite-sample convergence theorem. Coupling and
Gromov--Wasserstein background is standard
\cite{memoli2011gromov,villani2009optimal,peyre2019computational}.

\begin{definition}[Full-support metric--measure state]
A full-support metric--measure state is a tuple
$M=(X,d,\mu,\lambda)$, where $(X,d)$ is a finite metric space,
$\mu:X\to(0,1]$ satisfies $\sum_{x\in X}\mu(x)=1$, and
$\lambda:X\to\mathcal L$ is a label map. Full support means
$\mu(x)>0$ for every $x\in X$.
\end{definition}

Full support is required for the exactness result: a zero-mass entity could be
added or moved without changing any coupling objective. The measure is part of
the state, while coordinates or unlisted attributes are not implicitly treated
as geometric evidence.

\begin{definition}[Label-compatible coupling]
For full-support states $M_X=(X,d_X,\mu_X,\lambda_X)$ and
$M_Y=(Y,d_Y,\mu_Y,\lambda_Y)$, a coupling is a function
$\pi:X\times Y\to\R_{\geq0}$ with
\[
 \sum_{y\in Y}\pi(x,y)=\mu_X(x),
 \qquad
 \sum_{x\in X}\pi(x,y)=\mu_Y(y).
\]
It is label-compatible when
$\pi(x,y)>0\Rightarrow\lambda_X(x)=\lambda_Y(y)$.
\end{definition}

\begin{definition}[$p$-metric--measure discrepancy]
For $1\leq p<\infty$ and a label-compatible coupling $\pi$, put
\[
 \mathcal D_p(\pi)^p
 =
 \sum_{x,x'\in X}\sum_{y,y'\in Y}
 |d_X(x,x')-d_Y(y,y')|^p
 \pi(x,y)\pi(x',y').
\]
The metric--measure discrepancy is
\[
 \Delta_{\mathrm{mm},p}(M_X,M_Y)
 =\inf_{\pi}\mathcal D_p(\pi),
\]
where the infimum ranges over label-compatible couplings. If no such coupling
exists, the discrepancy is $+\infty$.
\end{definition}

This is a finite hard-label-constrained Gromov--Wasserstein-type objective
\cite{memoli2011gromov}.
It does not include coordinate costs, soft attribute costs, or a learned
correspondence. Joint feature--structure costs, for example, lead to fused
Gromov--Wasserstein formulations \cite{vayer2020fused}; such additions define a
different discrepancy and require their own identifiability analysis.

\begin{theorem}[Zero metric--measure discrepancy is exact]
Let $M_X$ and $M_Y$ be full-support metric--measure states. For every
$1\leq p<\infty$,
$\Delta_{\mathrm{mm},p}(M_X,M_Y)=0$ if and only if there is a bijection
$\varphi:X\to Y$ such that, for every $x,x'\in X$,
\[
 d_Y(\varphi(x),\varphi(x'))=d_X(x,x'),\qquad
 \lambda_Y(\varphi(x))=\lambda_X(x),\qquad
 \mu_Y(\varphi(x))=\mu_X(x).
\]
\end{theorem}

\begin{proof}
Suppose first that such a bijection exists. Define
\[
 \pi_\varphi(x,y)=
 \begin{cases}
  \mu_X(x),&y=\varphi(x),\\
  0,&\text{otherwise}.
 \end{cases}
\]
The first marginal is $\mu_X$. For $y=\varphi(x)$, bijectivity and
measure preservation make the second marginal equal to
$\mu_X(x)=\mu_Y(y)$; hence this is a coupling. Label preservation makes
it label-compatible. A positive product
$\pi_\varphi(x,y)\pi_\varphi(x',y')$ can occur only with
$y=\varphi(x)$ and $y'=\varphi(x')$, where the distance difference is
zero. Thus $\mathcal D_p(\pi_\varphi)=0$ and the discrepancy is zero.

Conversely, suppose the discrepancy is zero. The set of all couplings is a
closed and bounded subset of the finite-dimensional space
$\R^{X\times Y}$; imposing the finitely many label constraints keeps a
closed subset. If it is nonempty, it is compact. Since $\mathcal D_p$ is
continuous, a minimizing coupling $\pi_*$ exists and satisfies
$\mathcal D_p(\pi_*)=0$.

Let $C=\{(x,y):\pi_*(x,y)>0\}$. Full support of each marginal implies
that every $x$ occurs in some pair in $C$ and every $y$ occurs in some pair in
$C$, so $C$ is a correspondence. For any two pairs
$(x,y),(x',y')\in C$, their coefficient in the defining sum is positive.
Every summand is nonnegative and the sum is zero, so
$d_X(x,x')=d_Y(y,y')$.

If both $(x,y)$ and $(x,y')$ belong to $C$, applying this equality to the two
pairs gives $d_Y(y,y')=d_X(x,x)=0$, hence $y=y'$. If both $(x,y)$ and
$(x',y)$ belong to $C$, the same argument gives $d_X(x,x')=0$, hence
$x=x'$. Therefore $C$ is the graph of a bijection $\varphi:X\to Y$.
The distance equality makes $\varphi$ an isometry, and positive support
makes it label-preserving. Finally, the first marginal gives
$\mu_X(x)=\pi_*(x,\varphi(x))$, while the graph property makes the
second marginal give
$\mu_Y(\varphi(x))=\pi_*(x,\varphi(x))$. Thus the bijection is
measure-preserving.
\end{proof}

The theorem gives a precise zero-set interpretation for the metric--measure
layer. It does not prove robustness under sampling, rates of empirical
convergence, existence of a good coupling for arbitrary predicted carriers,
or superiority over coordinate losses. These are separate empirical claims.

\begin{proposition}[Label-compatible coupling feasibility]
A label-compatible coupling exists if and only if the total mass of every
label agrees:
\[
\sum_{x:\lambda_X(x)=\ell}\mu_X(x)
=
\sum_{y:\lambda_Y(y)=\ell}\mu_Y(y)
\qquad(\ell\in\mathcal L).
\]
\end{proposition}

\begin{proof}
Necessity is the preceding marginal argument. For sufficiency, for each label
$\ell$ choose the product coupling within its two label blocks, normalized by
their common mass. Summing these blockwise couplings gives a nonnegative matrix
with the required marginals and no cross-label mass. If a common label mass is
zero, its block is omitted.
\end{proof}

Define the tagged label maps by $\bar\lambda(x)=(A,\lambda_A(x))$ for
$x\in F(A)$ and $\bar\nu(y)=(A,\nu_A(y))$ for $y\in G(A)$.

\begin{corollary}[State-level metric--measure invariance]
If $\Sigma\cong_{\mathrm{str}}\Sigma'$ then the induced full-support
metric--measure states $(|F|,d,\mu,\bar\lambda)$ and
$(|G|,\rho,\eta,\bar\nu)$ have zero metric--measure discrepancy for every
$1\leq p<\infty$.
\end{corollary}

\begin{proof}
Use the structural isomorphism as the measure-preserving,
label-preserving isometry in the zero metric--measure discrepancy theorem.
\end{proof}

\begin{proposition}[Extended pseudometric property of the hard-label discrepancy]
On full-support finite metric--measure states with tagged labels,
$\Delta_{\mathrm{mm},1}$ is symmetric, vanishes on the diagonal, and satisfies
the triangle inequality. It may take the value $+\infty$ when a
label-compatible coupling does not exist.
\end{proposition}

\begin{proof}
The identity-graph coupling gives diagonal vanishing. Transposing any
label-compatible coupling preserves its marginals in reverse order and leaves
$\mathcal D_1$ unchanged, which proves symmetry after taking infima. For the
triangle inequality, let $\pi_{XY}$ and $\pi_{YZ}$ be label-compatible
couplings. Full support
gives $\mu_Y(y)>0$, so
\[
 \gamma(x,y,z)=\frac{\pi_{XY}(x,y)\pi_{YZ}(y,z)}{\mu_Y(y)}
\]
is well-defined. This is the standard gluing construction in optimal transport
\cite{villani2009optimal,peyre2019computational}. Direct summation shows that its $(X,Y)$ and
$(Y,Z)$ marginals are the original couplings. Let $\pi_{XZ}$ be its $(X,Z)$
marginal.
Whenever $\gamma(x,y,z)>0$, label compatibility gives
$\lambda_X(x)=\lambda_Y(y)=\lambda_Z(z)$; hence $\pi_{XZ}$ is
label-compatible. For two triples sampled from $\gamma$,
\[
 |d_X(x,x')-d_Z(z,z')|
 \leq |d_X(x,x')-d_Y(y,y')|+|d_Y(y,y')-d_Z(z,z')|.
\]
Summing this inequality against $\gamma\otimes\gamma$ yields
$\mathcal D_1(\pi_{XZ})\leq\mathcal D_1(\pi_{XY})+
\mathcal D_1(\pi_{YZ})$. Taking infima over the two original couplings proves
the triangle inequality. If either coupling is infeasible, the corresponding
extended value is $+\infty$ and the inequality is immediate.
\end{proof}

\section{Categorical Descent Extension}

The relational carrier in the formal framework is a functor on a schema
category. This section makes the descent condition explicit when the schema is
given by generators and path equations. The purpose is to prevent a list of
primitive relations from being mistaken for a well-defined categorical state.
Schemas as categories and graph-and-equation presentations are standard in
functorial data models \cite{spivak2012functorial,
spivakwisnesky2015relational}; TSI changes the instance codomain from sets and
functions to finite sets and binary relations.

\begin{definition}[Generator realization]
A generator realization assigns a finite nonempty set $X_A$ to every
$A\in Q_0$ and a relation $R_e\subseteq X_{s(e)}\times X_{t(e)}$ to
every $e\in Q_1$. Its free relational extension is
\[
 \widetilde F(1_A)=\Delta_{X_A},\qquad
 \widetilde F(e_n\cdots e_1)
 =R_{e_n}\circ\cdots\circ R_{e_1}.
\]
\end{definition}

\begin{proposition}[The finite-relation category]
Finite sets, binary relations, relational composition, and diagonal relations
form a category. In particular, the path formula in the preceding definition
is independent of parenthesization. This is the standard category of relations
\cite{freydscedrov1990categories}; the elementary verification is included for
completeness.
\end{proposition}

\begin{proof}
Let $R:X\nrightarrow Y$, $S:Y\nrightarrow Z$, and $T:Z\nrightarrow W$.
A pair $(x,w)$ belongs to $T\circ(S\circ R)$ exactly when there are
$y\in Y$ and $z\in Z$ with
$(x,y)\in R$, $(y,z)\in S$, and $(z,w)\in T$. Expanding
$(T\circ S)\circ R$ gives exactly the same existential condition, so
composition is associative. A pair belongs to $\Delta_Y\circ R$ exactly
when its intermediate element is forced by the diagonal to equal the target;
hence $\Delta_Y\circ R=R$. The analogous argument gives
$R\circ\Delta_X=R$. These are the category axioms.
\end{proof}

\begin{theorem}[Descent through path equations]
A generator realization descends to a functor
$F:\mathcal S\to\FinRel$ satisfying
$F([e])=R_e$ if and only if
\[
 \widetilde F(p)=\widetilde F(q)
 \qquad\text{for every }(p,q)\in\mathcal E.
\]
When it exists, the descended functor is unique.
\end{theorem}

\begin{proof}
Assume first that a descended functor $F$ exists and satisfies
$F([e])=R_e$. Let
$\pi:\operatorname{Path}(Q)\to\mathcal S$ map each path to its
equivalence class. We first verify the factorization from the generator data.
For a length-zero path, functoriality gives
$F([1_A])=\Delta_{X_A}=\widetilde F(1_A)$. If
$p=e_n\cdots e_1$ and the equality holds for paths of length $n-1$, then
functoriality and $F([e_i])=R_{e_i}$ give
\[
 F([p])=F([e_n])\circ F([e_{n-1}\cdots e_1])
 =R_{e_n}\circ\widetilde F(e_{n-1}\cdots e_1)=\widetilde F(p).
\]
Thus $\widetilde F=F\circ\pi$ by induction on path length. If
$(p,q)\in\mathcal E$, then
$\pi(p)=\pi(q)$ by construction. Therefore
$\widetilde F(p)=F(\pi(p))=F(\pi(q))=\widetilde F(q)$.

Conversely, assume equality for every declared equation. Define
$p\sim q$ exactly when $\widetilde F(p)=\widetilde F(q)$, for parallel
paths. Equality is reflexive, symmetric, and transitive. If $p\sim q$ and
$a,b$ are compatibly composable paths, functoriality of the free extension and
associativity give
\[
 \widetilde F(bpa)
 =\widetilde F(b)\circ\widetilde F(p)\circ\widetilde F(a)
 =\widetilde F(b)\circ\widetilde F(q)\circ\widetilde F(a)
 =\widetilde F(bqa).
\]
Thus $\sim$ is a path congruence containing $\mathcal E$. By minimality,
$\equiv_{\mathcal E}\subseteq\sim$. Hence
$F([p])=\widetilde F(p)$ is well-defined. The identity and composition laws
are inherited from $\widetilde F$, so $F$ is a functor and
$\widetilde F=F\circ\pi$. Every morphism of $\mathcal S$ has a path
representative, so any functor satisfying the factorization is forced to have
$F([p])=\widetilde F(p)$; uniqueness follows.
\end{proof}

\begin{proposition}[Generator criterion for natural equivalence]
Let $F,G:\mathcal S\to\FinRel$ be functors with the same object
carriers and let $\varphi_A:F(A)\to G(A)$ be bijections.
The graphs $\operatorname{Gr}(\varphi_A)$ form a natural isomorphism if and
only if, for every generator $e:A\to B$ and every $x,y$,
\[
 (x,y)\in F([e])
 \Longleftrightarrow
 (\varphi_A(x),\varphi_B(y))\in G([e]).
\]
\end{proposition}

\begin{proof}
If the family is natural, its naturality square at each generator is
$G([e])\circ\operatorname{Gr}(\varphi_A)
=\operatorname{Gr}(\varphi_B)\circ F([e])$. Expanding graph
composition gives the displayed biconditional: one direction follows by
transporting a related pair, and the other uses surjectivity of
$\varphi_B$.

Conversely, the biconditional transports each generator relation. It also
holds for identities because a bijection transports a diagonal relation to a
diagonal relation. For composable paths, membership in a relational composite
is witnessed by an intermediate entity; applying the criterion to both factors
and using bijectivity transports that witness. Induction on path length gives
the criterion for every path. Since both functors satisfy the same path
equations, it is independent of the chosen representative. The corresponding
naturality equality therefore holds for every schema morphism. Each graph is
an isomorphism in $\FinRel$, so the natural transformation is a natural
isomorphism.
\end{proof}

If a declared path equation is not satisfied by the generator relations, the
assignment remains a valid functor on the free path category but is not a
state over the presented schema. Thus categorical validity is an additional
mathematical gate, not an interpretive label attached after prediction.
The extension does not assert that a learner discovers the correct schema or
that generator-level agreement is sufficient without checking the equations.

\section{Dynamics and Temporal Identity Extension}

A structural transition is a partial identity claim: it says which typed
entities persist across an update. This section makes finite action histories
and turnover-aware defects explicit. Endpoint agreement alone is not treated
as evidence for temporal identity.

\begin{definition}[Finite action-prefix category]
Let $\mathcal A$ be a nonempty finite action alphabet and let $H\geq0$. The
category $\mathcal H_{\mathcal A}^H$ has as objects all words
$u\in\mathcal A^*$ with $|u|\leq H$. There is one morphism $u\to v$ exactly
when $u$ is a prefix of $v$; its identity is the empty extension and composition
concatenates suffixes.
\end{definition}

The prefix relation is a finite poset category. A morphism $u\to v$ represents
executing the suffix of $v$ that remains after $u$.

\begin{definition}[Tracked action history]
A tracked action history of horizon $H$ is a functor
$\mathcal D:\mathcal H_{\mathcal A}^H\to\StrTrack(Q,\mathcal E)$.
Thus $\mathcal D(u)$ is the state at action word $u$, while
$\mathcal D(u\to v)$ is the composite tracked transition from the prefix state
to the descendant state.
\end{definition}

\begin{theorem}[Unique extension from one-step action edges]
Suppose a state $\Sigma_u$ is assigned to every word with $|u|\leq H$ and a
tracked transition $p_{u,a}:\Sigma_u\rightharpoonup\Sigma_{ua}$ is assigned
whenever $|u|<H$. There is a unique functor $\mathcal D$ taking these one-step
transitions as prescribed values.
\end{theorem}

\begin{proof}
Set $\mathcal D(u)=\Sigma_u$ and map every identity to the identity transition.
If $v=ua_1\cdots a_k$, define
\[
\mathcal D(u\to v)=
p_{ua_1\cdots a_{k-1},a_k}\circ\cdots\circ p_{u,a_1}.
\]
Associativity gives independence of parenthesization, and splitting an edge list
at any intermediate prefix gives functoriality. The empty list gives identities.
Any functor with the prescribed one-step values must send every longer prefix
morphism to this composite, proving uniqueness.
\end{proof}

\begin{theorem}[Trajectory preservation along action histories]
Let $u\preceq v$ in a tracked action history. If every one-step edge on the path
from $u$ to $v$ preserves any chosen subset of topology, geometry, relations,
filtration, and mass on survivors, then $\mathcal D(u\to v)$ preserves the same
subset.
\end{theorem}

\begin{proof}
The unique extension theorem expresses $\mathcal D(u\to v)$ as the composite of
the one-step edges. The five-layer composition theorem applies successively.
\end{proof}

\begin{definition}[Turnover and layer defects]
For $p:\Sigma\rightharpoonup\Sigma'$ with
$\Sigma=(F,\lambda,K,d,f,\mu)$ and
$\Sigma'=(G,\nu,L,\rho,g,\eta)$, let
\[
b(p)=\sum_A(|F(A)|-|\dom(p_A)|+|G(A)|-|\im(p_A)|).
\]
Define
\begin{align*}
m_K(p)&=|\bar p[K|_{D_p}]\mathbin{\triangle}L|_{I_p}|,\\
m_d(p)&=\max_{x,y\in D_p}|d(x,y)-\rho(\bar p x,\bar p y)|,\\
m_R(p)&=\sum_{e\in Q_1}|(p_{s(e)}\times p_{t(e)})
 [F([e])|_{D_{s(e)},D_{t(e)}}]\mathbin{\triangle}
 G([e])|_{I_{s(e)},I_{t(e)}}|,\\
m_f(p)&=\max_{\sigma\in K|_{D_p}}
\begin{cases}
|f(\sigma)-g(\bar p\sigma)|,&\bar p\sigma\in L,\\
+\infty,&\bar p\sigma\notin L,
\end{cases}\\
m_\mu(p)&=\sum_{x\in D_p}|\mu(x)-\eta(\bar p x)|.
\end{align*}
If $D_p$ is empty, set $m_d(p)=m_f(p)=m_\mu(p)=0$. Here
$D_A=\dom(p_A)$ and $I_A=\im(p_A)$. The $+\infty$ convention makes the
filtration defect defined even when a surviving source simplex is not mapped to
a target simplex.
\end{definition}

For $e:A\to B$ and tagged subsets $D_A\subseteq F(A)$ and
$D_B\subseteq F(B)$, define the ordered simplicial-pair capacity
\[
c_e(K;D_A,D_B)=
\bigl|\{(x,y)\in D_A\times D_B:\{x,y\}\in K\}\bigr|.
\]
This convention includes a permitted loop $(x,x)$ through the vertex
$\{x\}\in K$ and therefore does not silently assume irreflexive relations.

\begin{proposition}[Compatibility bounds the relational defect]
For every tracked transition $p:\Sigma\rightharpoonup\Sigma'$ between
compatible states,
\[
m_R(p)\leq
\sum_{e\in Q_1}\bigl(c_e(K;D_{s(e)},D_{t(e)})
+c_e(L;I_{s(e)},I_{t(e)})\bigr).
\]
Without the compatibility axiom, this bound need not hold.
The bound is a capacity bound and is not claimed to be tight.
\end{proposition}

\begin{proof}
For each generator $e$, compatibility puts the restricted source relation
inside the source simplicial-pair set counted by $c_e(K;D_{s(e)},D_{t(e)})$.
Injectivity of $p_{s(e)}\times p_{t(e)}$ preserves its cardinality. Target
compatibility likewise puts the restricted target relation inside the set
counted by $c_e(L;I_{s(e)},I_{t(e)})$. The cardinality of a symmetric
difference is at most the sum of the cardinalities of its two operands; summing
over generators proves the bound.

For an explicit failure without compatibility, let $Q$ have distinct objects
$A,B$ and one generator $e:A\to B$. Take $F(A)=\{x\}$,
$F(B)=\{y\}$, $G(A)=\{x'\}$, and $G(B)=\{y'\}$, with the displayed
elements carrying their respective distinct type tags. Let $K$ and $L$ contain
only their vertices, set $F([e])=\{(x,y)\}$ and $G([e])=\varnothing$, and let
the total tracked transition map $x\mapsto x'$ and $y\mapsto y'$. Then both
capacities are zero, while
\[
m_R(p)=|\{(x',y')\}\mathbin{\triangle}\varnothing|=1.
\]
Thus the bound can fail when the source compatibility axiom is removed.
\end{proof}

\begin{proposition}[Zero criterion for a tracked transition]
For every tracked transition $p$, $m_K(p)=0$, $m_d(p)=0$, $m_R(p)=0$,
$m_f(p)=0$, and $m_\mu(p)=0$ are respectively equivalent to preservation of
topology, geometry, generator relations, filtration, and mass on survivors.
Also, $b(p)=0$ is equivalent to every component being a total bijection.
Consequently, all six quantities vanish exactly when $p$ is a full integrated
structural isomorphism.
\end{proposition}

\begin{proof}
All five layer statements follow because finite symmetric differences, maxima of nonnegative terms, and sums of nonnegative terms vanish
exactly when all their component conditions hold. The $+\infty$ case ensures
that $m_f(p)=0$ implies that every surviving simplex has a simplex image. For
$b(p)$, vanishing forces both the domain and image of every component to equal
the full finite carrier; injectivity then gives a total bijection. Combining
these facts with the label-preserving definition of tracked transition gives all
six conditions of integrated structural isomorphism. The converse follows by
substitution.
\end{proof}

\begin{proposition}[Survivor preservation can be vacuous]
For any two nonempty integrated states, the transition with empty domain and
empty image preserves topology, geometry, relations, filtration, and mass on
survivors, while having strictly positive turnover.
\end{proposition}

\begin{proof}
All induced survivor structures and all quantified comparisons are empty, while
nonempty endpoint carriers make the turnover sum positive.
\end{proof}

The dynamics extension therefore separates local structural preservation,
temporal identity, and turnover. It does not claim that terminal state
correctness identifies the tracking map, that a formal action branch is a causal
counterfactual, or that a learner discovers the correct update mechanism.

\section{Dynamics and Rollout Stability}

Static structure alone does not make a transition model well-defined. We require
updates to respect structural equivalence before measuring their error.

Bisimulation metrics construct task-dependent behavioral distances for Markov
decision processes from reward and transition information
\cite{ferns2004metrics}. Lipschitz model analyses propagate one-step error over
multiple steps in Wasserstein geometry \cite{asadi2018lipschitz}. The
discrepancy below is neither asserted to be a bisimulation fixed point nor
derived from rewards. The rollout theorem is instead conditional on a supplied
structural pseudometric and supplied Lipschitz constants.

\begin{definition}[Extended structural pseudometric]
Let $\mathcal X=\StrState(Q,\mathcal E)/\cong_{\mathrm{str}}$. An extended
structural pseudometric on a subset $\mathcal X_0\subseteq\mathcal X$ is a
map
\[
\Dstr:\mathcal X_0\times\mathcal X_0\to[0,\infty]
\]
that is symmetric, vanishes on the diagonal, satisfies the triangle
inequality, and is allowed to take the value $+\infty$. A finite-distance
component is a subset on which all pairwise values are finite. The rollout
theorem below is conditional on these properties; it does not claim that any
particular layer discrepancy automatically has them.
\end{definition}

\begin{proposition}[Existence of an admissible discrepancy]
For every subset $\mathcal X_0$, the discrete discrepancy
\[
\Dstr^{\mathrm{disc}}(x,y)=
\begin{cases}0,&x=y,\\1,&x\ne y,\end{cases}
\]
is a structural metric on $\mathcal X_0$. Every map
$T:\mathcal X_0\to\mathcal X_0$ is $1$-Lipschitz under this metric.
\end{proposition}

\begin{proof}
Symmetry and diagonal vanishing are immediate. For the triangle inequality,
if $x=z$ the result is trivial; otherwise at least one of $x\ne y$ or
$y\ne z$ holds, so $\Dstr^{\mathrm{disc}}(x,z)=1\leq
\Dstr^{\mathrm{disc}}(x,y)+\Dstr^{\mathrm{disc}}(y,z)$. A map cannot increase
a zero distance, and the only positive input distance is $1$, so the
Lipschitz claim follows.
\end{proof}

The discrete example is intentionally degenerate for rollout: if the one-step
tolerance is less than $1$, the predicted and true classes must coincide; if it
is at least $1$, the resulting bound can be vacuous because the diameter is
$1$.

The metric--measure result gives a non-discrete admissible instance. For a
state $\Sigma$, let $M(\Sigma)$ be its induced full-support metric--measure
state with tagged labels, and define
\[
 \Dstr^{\mathrm{mm}}([\Sigma],[\Sigma'])
 =\Delta_{\mathrm{mm},1}(M(\Sigma),M(\Sigma')).
\]
Structural isomorphisms induce label- and measure-preserving isometries, so the
value is independent of the chosen representatives. Symmetry, diagonal
vanishing, and the triangle inequality make this an extended structural
pseudometric on $\mathcal X$. It is not a metric there: zero value identifies
the metric--measure layer but need not identify the simplicial complex or the
generator relations.

\begin{assumption}[Quotient-defined Lipschitz action models]
Let $\mathcal X_0$ be a finite-distance component invariant under true and
predicted updates. For each action $a$, assume that the updates
$T_a,\widehat T_a:\mathcal X_0\to\mathcal X_0$ are well-defined, that $T_a$ is
$L_a$-Lipschitz under an extended structural pseudometric $\Dstr$, and that
$\Dstr(\widehat T_a z,T_a z)\leq\varepsilon_a$ at every predicted-rollout
state $z$ where the estimate is invoked.
\end{assumption}

The well-definedness condition is an equivariance assumption: isomorphic input
representatives must produce isomorphic outputs. It does not follow from the
state definition. A pseudometric is sufficient for the recurrence; a genuine
metric is required only when distinct quotient classes must be separated.

\begin{theorem}[Action-conditioned rollout bound]
For actions $a_1,\ldots,a_n$ and common initial class $x_0$, define
$x_i=T_{a_i}x_{i-1}$ and $\widehat x_i=\widehat T_{a_i}\widehat x_{i-1}$. Under
the assumption above,
\[
\Dstr(\widehat x_n,x_n)\leq
\sum_{i=1}^n\varepsilon_{a_i}\prod_{j=i+1}^nL_{a_j},
\]
where an empty product equals $1$.
\end{theorem}

\begin{proof}
Let $e_i=\Dstr(\widehat x_i,x_i)$. The triangle inequality gives
\[
e_i\leq\Dstr(\widehat T_{a_i}\widehat x_{i-1},T_{a_i}\widehat x_{i-1})+
\Dstr(T_{a_i}\widehat x_{i-1},T_{a_i}x_{i-1})\leq\varepsilon_{a_i}+L_{a_i}e_{i-1}.
\]
Since $e_0=0$, induction on $i$ expands this recurrence into the displayed
sum. Each induction step multiplies all previous terms by $L_{a_i}$ and adds
$\varepsilon_{a_i}$, which is exactly the claimed product pattern.
\end{proof}

\begin{proposition}[No uniform rollout bound without uniform Lipschitz control]
A common one-step error bound does not imply a nontrivial multi-step bound
uniform over a family of models whose Lipschitz constants are unbounded.
\end{proposition}

\begin{proof}
Take the metric component $[0,1]$ with discrepancy
$\Dstr(u,v)=c|u-v|$. For $\eta>0$, let
$T_M(u)=\min\{1,Mu\}$ and $\widehat T_M(u)=\min\{1,Mu+\eta\}$.
Both maps are well-defined on this component, and $T_M$ is $M$-Lipschitz.
The one-step discrepancy is at most $c\eta$, independently of $M$. Starting
at zero, choose $M$ so that $M\eta+\eta\geq1$; the predicted state reaches one
at step two while the true state remains zero. Thus a fixed two-step error
survives as $\eta\to0$ when no common Lipschitz bound is imposed.
\end{proof}

\begin{proposition}[Quotient equivariance is independently necessary]
A representative-level update that maps isomorphic input states to
non-isomorphic output states does not induce a well-defined update on
$\mathcal X=\StrState(Q,\mathcal E)/\cong_{\mathrm{str}}$.
\end{proposition}

\begin{proof}
Let $\Sigma\cong_{\mathrm{str}}\Sigma'$ and suppose a representative-level rule
sends them to $\Theta$ and $\Theta'$ with
$\Theta\not\cong_{\mathrm{str}}\Theta'$. The quotient class of the input is the
same whether represented by $\Sigma$ or $\Sigma'$, but the proposed output
class differs. Therefore no function on input equivalence classes can agree
with this rule on both representatives.
\end{proof}

\section{Interpretation and Limits}

The formal state has five coupled structural observables and a temporal
transition layer. Topology concerns qualitative organization such as adjacency
and connectedness. Geometry concerns metric information. Relations concern
typed entities, generator morphisms, and composition. Filtration and mass add
ordered and measure-theoretic observables on the same carrier. Dynamics concerns
action-conditioned tracked transitions, histories, turnover, interventions, and
rollouts. The common tagged carrier and the compatibility axiom couple these
layers without identifying them.

The formal results establish conditional statements about explicitly specified
objects. They do not prove that these observables are necessary or sufficient
for intelligence, that human cognition literally uses this state object, or
that an unconstrained neural network will discover it. They also do not prove
that a structurally specified model is universally superior. Those statements
require separate cognitive and empirical validation.

The principal design boundary is therefore deliberate: a structural claim must
name its carrier, equivalence, observables, transformations, and discrepancy
before it can be evaluated. A scalar predictive loss may be useful, but it
cannot by itself determine structural equivalence or transition preservation.
Conversely, a structural discrepancy is meaningful only after its task-relevant
alignment and invariants are specified.

The formalism takes the entity set, its typing, and its cross-time identity as
given data of the state. It does not supply a procedure for discovering
entities from raw observation or for maintaining identity under perception
error. Those stages precede everything specified here.

The companion empirical papers test bounded consequences of this framework on
finite mechanism families. Those experiments should be read as validation of
the stated consequences, not as validation of the entire theory or of
real-world cognition.

\section{Conclusion}

This paper turns the TSI proposal into a formal core. An integrated structural
state is a typed relational functor on a finite presented schema, a simplicial
complex and metric on a common tagged carrier, a monotone filtration, a
full-support mass, and action-conditioned tracked transitions. The compatibility
axiom makes generator relation support a subcomplex of the simplicial
one-skeleton. Structural equivalence is an equivalence relation; preservation of
the five observables composes on survivors; the six defects characterize full
structural isomorphism together with turnover; and the quotient-level rollout
bound makes equivariance and stability assumptions explicit.

These theorems do not settle the cognitive or machine-learning hypotheses that
motivated TSI. They provide definitions, proofs, and failure boundaries needed
to state those hypotheses without leaving intuition as an untestable remainder.

The next decisive comparison is not a factorized head versus a generic sparse
head. It contrasts two models that have both recovered the same dependency
graph under entity creation and deletion, identity maintenance, and relation
reconfiguration. For that comparison to identify anything, both arms must
receive identical observable information and differ only in representational
commitment. Supplying identity or typed relations as input to one arm only
would make the result an artifact of information asymmetry, the same class of
confound diagnosed in the original finite-family audit. The turnover quantity
\(b(p)\), its zero criterion, and the vacuity proposition for survivor-only
preservation are already formalized here; a turnover-focused study is the
smallest self-contained instance of this design.

\section*{Acknowledgments}
We thank the colleagues and acquaintances who reviewed the ideas underlying
this study.

\section*{Data and Code Availability}
This theoretical study introduces no new empirical dataset. The bilingual
manuscript sources and reference implementation are available at
\url{https://github.com/nowtast/tsi}.
\bibliographystyle{plainnat}
\bibliography{references}

@inproceedings{battaglia2016interaction,
  title={Interaction Networks for Learning about Objects, Relations and Physics},
  author={Battaglia, Peter W. and Pascanu, Razvan and Lai, Matthew and Rezende, Danilo Jimenez and Kavukcuoglu, Koray},
  booktitle={Advances in Neural Information Processing Systems},
  volume={29},
  year={2016},
  url={https://proceedings.neurips.cc/paper/2016/hash/3147da8ab4a0437c15ef51a5cc7f2dc4-Abstract.html}
}

@article{battaglia2018relational,
  title={Relational Inductive Biases, Deep Learning, and Graph Networks},
  author={Battaglia, Peter W. and Hamrick, Jessica B. and Bapst, Victor and S{\'a}nchez-Gonz{\'a}lez, {\'A}lvaro and Zambaldi, Vinicius and Malinowski, Mateusz and Tacchetti, Andrea and Raposo, David and Santoro, Adam and Faulkner, Ryan and G{\"u}l{\c c}ehre, {\c C}a{\u g}lar and Song, Francis and Ballard, Andrew and Gilmer, Justin and Dahl, George and Vaswani, Ashish and Allen, Kelsey and Nash, Charles and Langston, Victoria and Dyer, Chris and Heess, Nicolas and Wierstra, Daan and Kohli, Pushmeet and Botvinick, Matt and Vinyals, Oriol and Li, Yujia and Pascanu, Razvan},
  journal={arXiv preprint arXiv:1806.01261},
  year={2018},
  doi={10.48550/arXiv.1806.01261}
}

@inproceedings{kipf2020contrastive,
  title={Contrastive Learning of Structured World Models},
  author={Kipf, Thomas and van der Pol, Elise and Welling, Max},
  booktitle={International Conference on Learning Representations},
  year={2020},
  url={https://openreview.net/forum?id=H1gax6VtDB}
}

@book{freydscedrov1990categories,
  title={Categories, Allegories},
  author={Freyd, Peter J. and Scedrov, Andre},
  series={North-Holland Mathematical Library},
  volume={39},
  publisher={North-Holland},
  address={Amsterdam},
  year={1990},
  isbn={978-0-444-70368-2}
}

@article{spivak2012functorial,
  title={Functorial Data Migration},
  author={Spivak, David I.},
  journal={Information and Computation},
  volume={217},
  pages={31--51},
  year={2012},
  doi={10.1016/j.ic.2012.05.001}
}

@inproceedings{spivakwisnesky2015relational,
  title={Relational Foundations for Functorial Data Migration},
  author={Spivak, David I. and Wisnesky, Ryan},
  booktitle={Proceedings of the 15th Symposium on Database Programming Languages},
  pages={21--28},
  publisher={ACM},
  year={2015},
  doi={10.1145/2815072.2815075}
}

@article{edelsbrunner2002topological,
  title={Topological Persistence and Simplification},
  author={Edelsbrunner, Herbert and Letscher, David and Zomorodian, Afra},
  journal={Discrete \& Computational Geometry},
  volume={28},
  number={4},
  pages={511--533},
  year={2002},
  doi={10.1007/s00454-002-2885-2}
}

@article{zomorodian2005computing,
  title={Computing Persistent Homology},
  author={Zomorodian, Afra and Carlsson, Gunnar},
  journal={Discrete \& Computational Geometry},
  volume={33},
  number={2},
  pages={249--274},
  year={2005},
  doi={10.1007/s00454-004-1146-y}
}

@article{cohensteiner2007stability,
  title={Stability of Persistence Diagrams},
  author={Cohen-Steiner, David and Edelsbrunner, Herbert and Harer, John},
  journal={Discrete \& Computational Geometry},
  volume={37},
  number={1},
  pages={103--120},
  year={2007},
  doi={10.1007/s00454-006-1276-5}
}

@inproceedings{chazal2009proximity,
  title={Proximity of Persistence Modules and Their Diagrams},
  author={Chazal, Fr{\'e}d{\'e}ric and Cohen-Steiner, David and Glisse, Marc and Guibas, Leonidas J. and Oudot, Steve Y.},
  booktitle={Proceedings of the 25th Annual Symposium on Computational Geometry},
  pages={237--246},
  publisher={ACM},
  year={2009},
  doi={10.1145/1542362.1542407}
}

@article{memoli2011gromov,
  title={{Gromov--Wasserstein} Distances and the Metric Approach to Object Matching},
  author={M{\'e}moli, Facundo},
  journal={Foundations of Computational Mathematics},
  volume={11},
  number={4},
  pages={417--487},
  year={2011},
  doi={10.1007/s10208-011-9093-5}
}

@article{vayer2020fused,
  title={Fused {Gromov--Wasserstein} Distance for Structured Objects: Theoretical Foundations and Mathematical Properties},
  author={Vayer, Titouan and Chapel, Laetitia and Flamary, R{\'e}mi and Tavenard, Romain and Courty, Nicolas},
  journal={Algorithms},
  volume={13},
  number={9},
  pages={212},
  year={2020},
  doi={10.3390/a13090212}
}

@article{peyre2019computational,
  title={Computational Optimal Transport: With Applications to Data Science},
  author={Peyr{\'e}, Gabriel and Cuturi, Marco},
  journal={Foundations and Trends in Machine Learning},
  volume={11},
  number={5--6},
  pages={355--607},
  year={2019},
  doi={10.1561/2200000073}
}

@inproceedings{ferns2004metrics,
  title={Metrics for Finite {Markov} Decision Processes},
  author={Ferns, Norm and Panangaden, Prakash and Precup, Doina},
  booktitle={Proceedings of the 20th Conference on Uncertainty in Artificial Intelligence},
  pages={162--169},
  publisher={AUAI Press},
  year={2004},
  doi={10.48550/arXiv.1207.4114}
}

@inproceedings{asadi2018lipschitz,
  title={{Lipschitz} Continuity in Model-Based Reinforcement Learning},
  author={Asadi, Kavosh and Misra, Dipendra and Littman, Michael L.},
  booktitle={Proceedings of the 35th International Conference on Machine Learning},
  series={Proceedings of Machine Learning Research},
  volume={80},
  pages={264--273},
  publisher={PMLR},
  year={2018},
  url={https://proceedings.mlr.press/v80/asadi18a.html}
}

@article{bauerlesnick2015induced,
  title={Induced Matchings and the Algebraic Stability of Persistence Barcodes},
  author={Bauer, Ulrich and Lesnick, Michael},
  journal={Journal of Computational Geometry},
  volume={6},
  number={2},
  pages={162--191},
  year={2015},
  doi={10.20382/jocg.v6i2a9}
}

@article{crawleyboevey2015decomposition,
  title={Decomposition of Pointwise Finite-Dimensional Persistence Modules},
  author={Crawley-Boevey, William},
  journal={Journal of Algebra and Its Applications},
  volume={14},
  number={5},
  pages={1550066},
  year={2015},
  doi={10.1142/S0219498815500668}
}

@book{villani2009optimal,
  title={Optimal Transport: Old and New},
  author={Villani, C{\'e}dric},
  series={Grundlehren der mathematischen Wissenschaften},
  volume={338},
  publisher={Springer},
  address={Berlin, Heidelberg},
  year={2009},
  doi={10.1007/978-3-540-71050-9}
}

@book{maclane1998categories,
  title={Categories for the Working Mathematician},
  author={Mac Lane, Saunders},
  edition={2},
  series={Graduate Texts in Mathematics},
  volume={5},
  publisher={Springer},
  address={New York},
  year={1998},
  doi={10.1007/978-1-4757-4721-8}
}

@inproceedings{janner2019trust,
  title={When to Trust Your Model: Model-Based Policy Optimization},
  author={Janner, Michael and Fu, Justin and Zhang, Marvin and Levine, Sergey},
  booktitle={Advances in Neural Information Processing Systems},
  volume={32},
  pages={12498--12509},
  year={2019},
  url={https://papers.nips.cc/paper_files/paper/2019/hash/5faf461eff3099671ad63c6f3f094f7f-Abstract.html}
}
\end{document}